\documentclass[runningheads]{llncs}
\usepackage[T1]{fontenc}
\usepackage{comment}% Added for comments
\usepackage{pgfplots}% Added for figures
\pgfplotsset{compat=1.16}
\usepackage{xurl}
\usepackage{xcolor}
\usepackage{booktabs}
\usepackage{cite}
\usepackage{hyperref}
\usepackage{amsmath}
\usepackage{algorithm}
\usepackage{algpseudocode}
\usepackage{subcaption}

\usepackage{graphicx,verbatim}
\begin{document}
\title{Optimization-Based Calibration for Intravascular Ultrasound Volume Reconstruction}
\titlerunning{Optimization-Based Calibration for IVUS Volume Reconstruction}

\author{
Karl-Philippe Beaudet$^*$\,$^\dagger$\inst{1,2,4} \and 
Sidaty El Hadramy$^*$\inst{1,2,3,4} \and 
Philippe C. Cattin\inst{3} \and 
Juan Verde\inst{1,2,4} \and 
Stéphane Cotin\inst{1,2}
}

\authorrunning{K.-P. Beaudet, S. El Hadramy et al.}

\institute{
Inria, Strasbourg, France \and 
University of Strasbourg, CNRS, INSERM, ICube, UMR7357, Strasbourg, France \and 
Department of Biomedical Engineering, University of Basel, Switzerland \and 
Institute of Image-Guided Surgery, IHU Strasbourg, Strasbourg, France
}

\maketitle

\def\thefootnote{*}
\footnotetext{These authors contributed equally to this work}
\def\thefootnote{$\dagger$}
\footnotetext{Corresponding author: \email{kp.beaudet@gmail.com}}
\def\thefootnote{\arabic{footnote}}

\begin{abstract}
Intraoperative ultrasound images are inherently challenging to interpret in liver surgery due to the limited field of view and complex anatomical structures. Bridging the gap between preoperative and intraoperative data is crucial for effective surgical guidance. 3D IntraVascular UltraSound (IVUS) offers a potential solution by enabling the reconstruction of the entire organ, which facilitates registration between preoperative computed tomography (CT) scans and intraoperative IVUS images. In this work, we propose an optimization-based calibration method using a 3D-printed phantom for accurate 3D Intravascular Ultrasound volume reconstruction. Our approach ensures precise alignment of tracked IVUS data with preoperative CT images, improving intraoperative navigation. We validated our method using \textit{in vivo} swine liver images, achieving a calibration error from $0.88$ to $1.80$~mm and a registration error from $3.40$ to $5.71$~mm between the 3D IVUS data and the corresponding CT scan. Our method provides a reliable and accurate means of calibration and volume reconstruction. It can be used to register intraoperative ultrasound images with preoperative CT images in the context of liver surgery, and enhance intraoperative guidance.

% Authors must provide keywords and are not allowed to remove this Keyword section.

\end{abstract}
\section{Introduction}\label{introduction}

IntraVascular UltraSound (IVUS) is a medical imaging technique that involves an ultrasound (US) probe mounted at the tip of a catheter, which can be navigated inside blood vessels. Although IVUS is commonly used in cardiovascular procedures \cite{Mintz1995-qo}, recent work shows that it also has meaningful applications in liver surgery, where its real-time, minimally invasive, and radiation-free imaging capabilities can offer substantial benefits for intraoperative guidance \cite{Urade2021-gj}. Reconstructing a 3D volume from 2D IVUS images helps visualize and/or compute spatial relationships between internal structures, facilitating its use in computer-assisted interventions \cite{PhDThesis,beaudet2024towards,Hadramy2024-xx}. This process includes estimating relative spatial transformations between 2D frames. Although numerous approaches have been proposed to find these transformations without any tracking system \cite{Prevost2017-bi,El_hadramy2023-zm,Luo2022-cg}, these methods often suffer from cumulative drift errors, limiting their effectiveness in applications that require high precision. A tracking system enables accurate drift-free 3D reconstruction of the US volume. However, when manually mounted on an ultrasound probe, the sensor position relative to the image coordinate system is unknown. Therefore, calibration is necessary to determine the transformation that maps the 2D pixel's coordinates in the US image to the 3D coordinates in the sensor reference frame.

% Urade et \textit{al.} \cite{Urade2021-gj} demonstrated that when the catheter tip is positioned in the lumen of the inferior vena cava, it is possible to navigate a large part of the liver by solely rotating the catheter, without the need for X-ray imaging. Providing a serie of 2D images of the liver and its internal structures.

%Unlike Laparoscopic Ultrasound (LUS) in minimally invasive interventions, IVUS provides continuous and synchronized imaging without occupying any incision space. Additionally, it does not deform the liver, avoiding complications in image interpretation \cite{}. 

Various methods for calibrating ultrasound probes have been proposed \cite{Mercier2005-bj}. They can generally be categorized into two main approaches. Image-based calibration \cite{Wein2008-rn} typically relies on image registration techniques to compare images of the same region from different views, estimating the spatial transformation that aligns them. These methods are advantageous because they do not require additional hardware. However, they are limited by factors such as image quality \cite{Ma2008-ds}, lack of distinct features, and the small field of view, especially in the case of IVUS \cite{Mintz1995-qo}. Another class of methods is tool-based calibration, which relies on external objects \cite{Shen2019-eu}. A common approach involves using a tracked stylus with a known position, placed at multiple locations so that its tip appears in the ultrasound image from different angles \cite{Mercier2005-bj}. %While this method is widely used today, it is complex because it requires an additional tracking system. 
Other methods rely on dedicated calibration phantoms with well-defined geometric properties, such as spheres, point targets, crossing wires, or more complex 3D-printed structures \cite{Ronchetti2022-iy,Wu2023-cx}. Positioning the phantom within the field of view of the ultrasound probe, the system correlates the known geometry with the acquired image data to determine the transformation. However, existing phantoms were designed for standard probes with less constrained in-plane motion, while IVUS operates with predominantly out-of-plane motion and a constrained environment \cite{Mintz1995-qo,PhDThesis}. Additionally, some methods restrict probe orientation, such as wire phantoms with limited space between walls, leading to under-sampling of the six degrees of freedom. This results in varying calibration accuracy between in-plane and out-of-plane motion, further limiting the applicability of the existing methods for IVUS calibration.

In this work, we propose a calibration method adapted to IVUS probes. Our method relies on the 3D design of a phantom that constrains the environment of the IVUS probe and allows for out-of-plane rotational motion, mimicking the conditions of an IVUS inserted into a blood vessel. Landmarks from the phantom are detected in the acquired images and used to find calibration parameters that maximize the accuracy of the 3D reconstructed volume. To the best of our knowledge, this is the first method to propose a calibration approach designed for an IVUS probe. Our contributions are as follows: \textbf{1)} A phantom designed for IVUS calibration, considering space constraints and rotational motion. \textbf{2)} A gradient-based optimization method for IVUS probe calibration. \textbf{3)}  Experiments on \textit{in-vivo} porcine liver demonstrating the method’s reliability and efficiency. 
%\textbf{4)} Integration of the calibration method into real-world IVUS-assisted procedures.

The paper is organized as follows: Section \ref{problem_statement} states the mathematical formulation of the problem, Section \ref{Methods} describes the proposed method and its novel aspects, Section \ref{experiments} presents the experiments conducted, highlighting the accuracy and efficiency of the proposed approach. Section \ref{conclusion} concludes this work and discusses potential directions for future work.

\section{Problem formulation}\label{problem_statement}

\begin{figure}[ht]
    \centering
    \includegraphics[width=\textwidth]{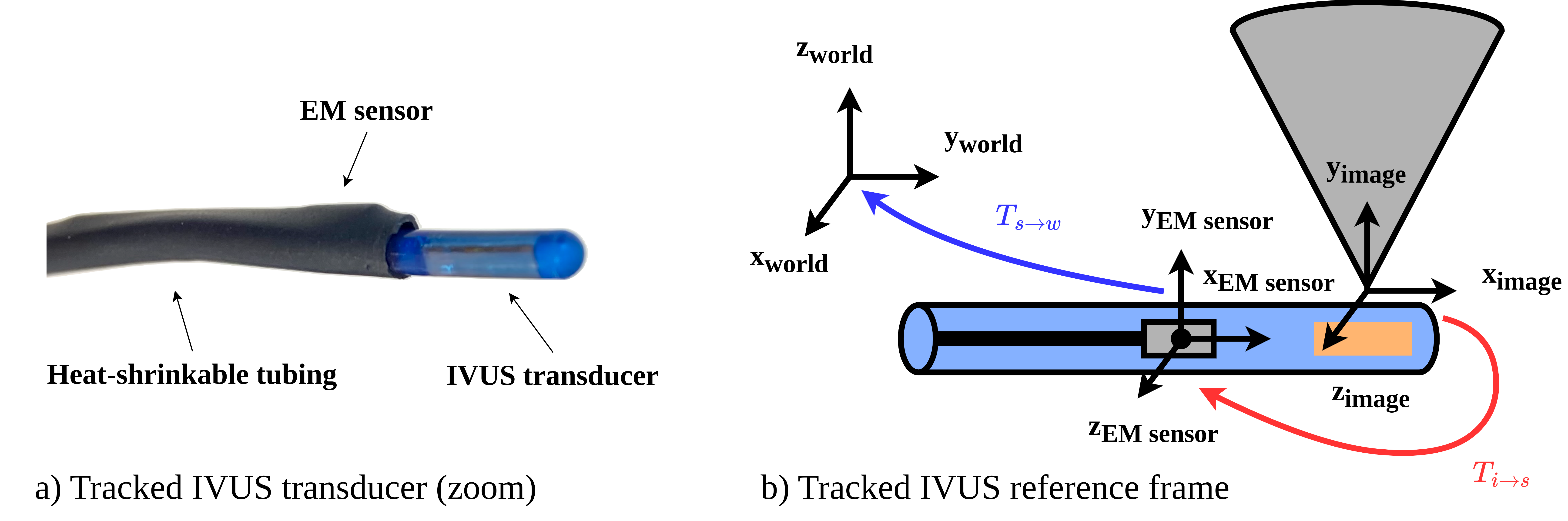}
    \caption{Illustration of the tracked IVUS setup. A sensor is mounted near the IVUS probe tip, capturing a 2D image. The figure shows three reference frames (world, sensor, and image) and the transformations: the known $T_{s \rightarrow w}$ and the unknown  $T_{i \rightarrow s}$, which includes translation, rotation, and scale. We consider here the world reference frame as the one where the sensor coordinates are expressed.}
    \label{fig:problem_statement}
\end{figure}

We address the spatial calibration problem for a sensor \textbf{manually} mounted near the tip of an IVUS probe, as illustrated in Figure~\ref{fig:problem_statement}. The probe captures 2D images within its field of view. Given the transformation $T_{s \rightarrow w}$ from the sensor to the world reference frame and the acquired images, the aim is to determine the unknown transformation $T_{i \rightarrow s}$ from the image to the sensor reference frame. This transformation is represented as a $4 \times 4$ matrix that includes translation, rotation, and scaling.

\section{Materials and methods} \label{Methods}
We propose a two-step approach: a 3D-printed phantom to constrain the IVUS probe and enable rotational motion, followed by an optimization loop to estimate the calibration matrix. While we use electromagnetic (EM) tracking in this work, the proposed method is adaptable to any system providing translation and orientation information. The following sections detail both steps.

%to obtain the transformation matrix from the image reference to the electromagnetic (EM) sensor reference attached to the IVUS. Given the nature of IVUS (catheter), we created a phantom 

\subsection{Phantom design}
\begin{figure}[ht]
    \centering
    \includegraphics[width=0.4\textwidth]{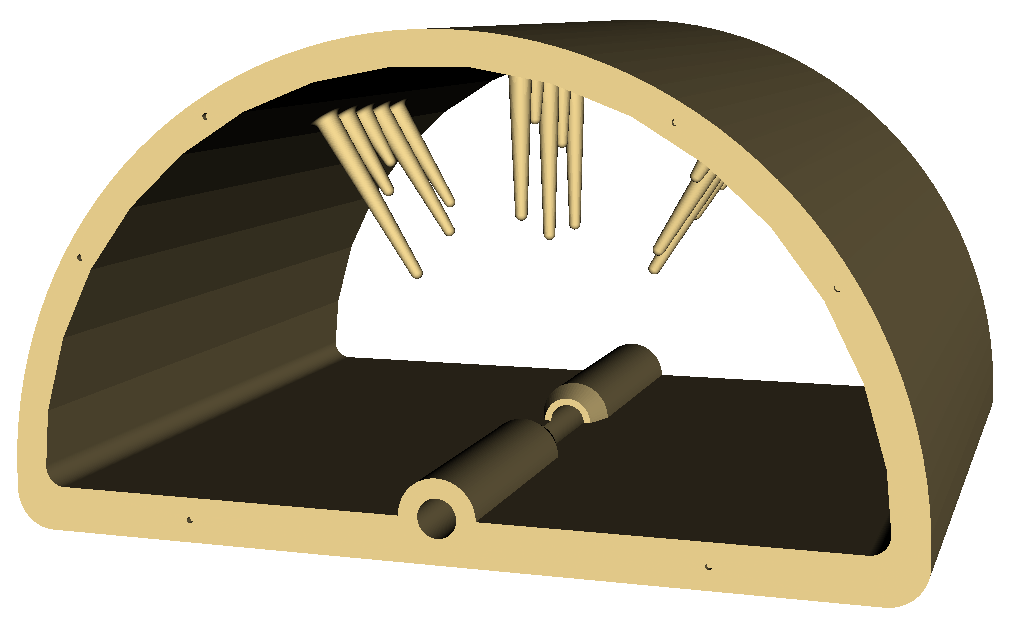}
    \caption{IVUS calibration phantom with a half-cylinder cavity and three marker clusters at 60°, 90°, and 120° defining distinct poses and proper image orientation.}
    \label{fig:phantom}
\end{figure}

We designed an IVUS-specific calibration phantom, illustrated in Figure~\ref{fig:phantom}, that constrains the probe within a tube, enabling controlled rotation and translation along one axis. The phantom incorporates needle-shaped markers arranged in three distinct clusters positioned at $60$°, $90$°, and $120$° to define three IVUS poses. Each cluster comprises five asymmetrically arranged needles aligned with the IVUS image plane to facilitate their identification in the image plane. Their lengths range from $1$ to $5$ cm. The phantom features a half-cylinder cavity with a $6.5$ cm internal radius, ensuring that all markers are visible at ultrasound depths ranging from $7$ to $11$ cm. %Designed for efficient 3D printing, 
The model was printed in PLA+ white with 15\% infill using tree supports. The process took 3 hours and 29 minutes on a Bambu Lab X1-Carbon 3D printer.

%We developed an IVUS-specific calibration phantom that constrains the probe within a tube, allowing controlled rotation and translation along one axis for easier manipulation. The phantom features needle-shaped markers designed to capture three distinct IVUS poses. Five asymmetrically arranged needles align with the IVUS image plane, ensuring proper orientation and preventing ambiguity. Their lengths range from $1$ to $5$ cm, ensuring the probe remains in-plane with the markers during acquisition. The phantom features a half-cylinder cavity with an internal radius of $6.5$ cm, allowing calibration at ultrasound depths between $6$ and $10$ cm. The three marker patterns are positioned at 60°, 90°, and 120° within the phantom. The model was designed for optimal 3D printing with supports. We 3D printed the model in PLA+ white with 15\% infill using tree supports. The printing process took 3 hours and 29 minutes with the Bambu Lab X1-Carbon 3D Printer.

%An electromagnetic (EM) sensor holder for the NDI TrackStar Model 800 was also incorporated into the phantom for tracking purposes. Additionally, 12 point markers were included to ensure needle-based spatial calibration from the phantom's EM sensor to its origin.

\subsection{Calibration process}

Using the IVUS-specific phantom described in the previous section, we address the calibration problem formulated in Section~\ref{problem_statement}. This is achieved by solving an optimization problem that finds the calibration matrix, ensuring a minimal discrepancy between the ultrasound-reconstructed volume of the phantom and its 3D model when both are registered. We propose a gradient-based optimization process, outlined in Algorithm~ \ref{alg:calibration_process}, where the parameters of the calibration matrix are iteratively adjusted to minimize a defined objective function. 
%The following sections detail each step of this process.

% The calibration procedure aims to align ultrasound image data with a phantom's ground truth points by iteratively optimizing the transformation parameters through a gradient-based approach. This process involves the definition of a calibration matrix, the projection of ultrasound image points, the registration of the transformed image points to the Phantom points, and the computation of the error between the two point clouds. In the following, the optimization algorithm is detailed in \ref{alg:calibration_process} and the detailed steps and associated mathematical formulations are provided.

\begin{algorithm}
\caption{Calibration matrix optimization process}\label{alg:calibration_process}
\begin{algorithmic}[0]
\State \textbf{Input:} The phantom needle points coordinates (P\textsubscript{Phantom}), the needle points coordinates in the ultrasound images (P\textsubscript{US}), the pose transforms of EM tracking sensor in which the needle points are visible (T\textsubscript{Sensor}), the error threshold (\( \epsilon \))
\State \textbf{Output:} An estimation of the calibration matrix \( C \)
\end{algorithmic}
\begin{algorithmic}[1]
\State \( \text{params} \leftarrow \text{(0,0,0,0,0,0,1)} \) \Comment{(\(R_x, R_y, R_z, t_x, t_y, t_z, s\))}
\State \( \text{optimizer} \leftarrow \text{Adam(params, lr)} \)
\State \( \text{error} \leftarrow \infty \)  
\While{\( \text{error} > \epsilon \)}
    \State \( C \leftarrow \text{calibrationMatrix(params)} \) \Comment{\textbf{Step 1:} Define Calibration matrix}
    \State \( P'\textsubscript{US} \leftarrow T\textsubscript{Sensor} \cdot C \cdot P\textsubscript{US} \) \Comment{\textbf{Step 2:} Apply calibration matrix}
    \State \( T\textsubscript{Registration} \leftarrow \text{ICP}(P'\textsubscript{US}, P\textsubscript{Phantom}) \) \Comment{\textbf{Step 3:} Compute ICP registration}
    \State \( P''\textsubscript{US} \leftarrow P'\textsubscript{US} \cdot T\textsubscript{Registration} \) \Comment{\textbf{Step 3:} Apply rigid registration}
    \State \( \text{error} \leftarrow \frac{1}{n} \sum_{i=1}^{n} \| P''\textsubscript{US}_i - P\textsubscript{Phantom}_i \|_2^2 \) \Comment{\textbf{Step 4:} Evaluate the cost function}
    \State \(\text{params} \leftarrow \text{optimizer(error)} \) \Comment{\textbf{Step 5:} Optimization}
\EndWhile
\State \( C \leftarrow \text{calibrationMatrix(params)} \) \Comment{Optimized calibration matrix definition}
\end{algorithmic}
\end{algorithm}

\textbf{Step 1: Calibration matrix definition}:
The calibration matrix, denoted $\mathbf{C}$, is defined with 7 parameters: 3 Euler angles (\(\text{roll}, \text{pitch}, \text{yaw}\)), 3 translations (\(t_x, t_y, t_z\)) and a scaling factor \(s\). It reads as: 
\begin{equation}
    \mathbf{C} = \begin{bmatrix}
    s \cdot \mathbf{R}_{\text{yaw}} \cdot \mathbf{R}_{\text{pitch}} \cdot \mathbf{R}_{\text{roll}} & [t_x, t_y, t_z]^T \\
    0 & 1
    \end{bmatrix}
    \label{eq:calibration_matrix_definition}
\end{equation}

%\noindent Here, \( \mathbf{R}_{\text{yaw}} \), \( \mathbf{R}_{\text{pitch}} \), and \( \mathbf{R}_{\text{roll}} \) are the rotation matrices corresponding to the Euler angles. The calibration matrix \( \mathbf{C} \) is the unknown we seek to determine. However, instead of optimizing directly over all 16 elements of \( \mathbf{C} \), we optimize over the Euler angles and translations. 
\noindent The calibration matrix \( \mathbf{C} \) is the unknown we seek to determine. We represent the rotation matrices using Euler angles and optimize over these angles and the translations. Additionally, using this approach, we guarantee that \( \mathbf{C} \) remains a calibration matrix at each iteration of the optimization process.

\textbf{Step 2: Projection of ultrasound image landmarks}:
At each iteration of the optimization process, the updated parameters are used to estimate the calibration matrix \( \mathbf{C} \) as described in the previous step. This matrix is then applied to project manually detected landmarks %from the ultrasound images 
from the image reference frame to the world reference frame using the equation: $\mathbf{p}_{\text{world}} =  \mathbf{T}_{\text{sensor}} \cdot \mathbf{C} \cdot \mathbf{p}_{\text{image}}$. Here, \( \mathbf{T}_{\text{sensor}} \) is the transformation matrix provided by the EM tracker, and \( \mathbf{p}_{\text{image}} = [x, y, 0, 1]^T \) represents the landmark positions in the image reference frame. These landmarks correspond to the needle tips, which can be easily identified and manually selected in the ultrasound images thanks to the phantom’s design choice. Thus, the landmarks are transferred to the world reference frame.

% Once the calibration matrix \( \mathbf{C} \) is defined, the next step is to project the ultrasound image points \( \mathbf{P}_{\text{image}} \) from the image reference frame into the world reference frame. This projection requires incorporating the pose of the electromagnetic sensor into the transformation. The final transformation is given by:

% \[
% \mathbf{p}_{\text{world}} = \mathbf{T}_{\text{Sensor}} \cdot \mathbf{C} \cdot \mathbf{p}_{\text{image}}
% \]

% Where \( \mathbf{T}_{\text{Sensor}} \) is the transformation matrix that represents the pose of the electromagnetic (EM) sensor in the world reference frame, \( \mathbf{C} \) is the calibration matrix defined in Step 1, and \( \mathbf{p}_{\text{ultrasound}} = [x, y, 0, 1]^T \) represents a point in the ultrasound image reference frame (where z = 0).

% The resulting point in the world reference frame is:

% \[
% \mathbf{p}_{\text{world}} = \begin{bmatrix} x' & y' & z' & 1 \end{bmatrix}^T
% \]

% This process ensures that ultrasound points are transformed consistently into the world reference frame, facilitating comparison with the ground truth phantom points.

\textbf{Step 3: ICP registration}:
At each optimization iteration, the given calibration matrix is used to transform the landmark positions into the world reference frame (\(\mathbf{p}_{\text{world}}\)). These transformed landmarks are then rigidly registered to their corresponding points in the phantom’s 3D model using point-to-point Iterative Closest Point (ICP) \cite{Besl1992-fp}. The ICP algorithm iteratively minimizes the alignment error, producing a registration matrix \( \mathbf{R}_{\text{ICP}} \) that ensures the best possible fit between the two landmark sets. We assume that any remaining misalignment after ICP reflects the error introduced by the calibration matrix at that iteration, providing a direct measure of its accuracy. 

% Next, the transformed ultrasound points are registered with the phantom points using the Iterative Closest Point (ICP) algorithm. The objective is to minimize the spatial discrepancy between the two point clouds: the transformed ultrasound points and the phantom's points.

% Let the ground truth points be denoted as \( \mathbf{P}_{\text{Phantom}} \), and the transformed ultrasound points as \( \mathbf{P}_{\text{transformed}} \). The ICP algorithm iteratively minimizes the following objective function:

% %\[
% %E_{\text{ICP}} = \sum_{i=1}^{N} \|\mathbf{p}_{\text{GT}, i} - \mathbf{R}_{\text{ICP}} \cdot \mathbf{T}_{\text{tracking}} \cdot \mathbf{C} \cdot \mathbf{p}_{\text{ultrasound}}\|^2
% %\]

% \[
% E_{\text{ICP}} = \sum_{i=1}^{N} \|\mathbf{p}_{\text{Phantom}, i} - \mathbf{R}_{\text{ICP}} \cdot \mathbf{p}_{\text{world, i}}\|^2
% \]

% Where \( N \) is the number of point correspondences, and \( \|\cdot\| \) represents the Euclidean distance between corresponding points in the two point clouds. This registration process yields a registration matrix \( \mathbf{R}_{\text{ICP}} \) that further refines the alignment of the ultrasound points with the ground truth.

\textbf{Step 4: Objective function evaluation}:
After the ICP registration, we evaluate the objective function defined in Equation \ref{eq:objective_function_evaluation} to measure the error introduced by the calibration matrix. This function quantifies the residual misalignment between the transformed ultrasound landmarks and their homologous in the phantom’s 3D model. The resulting error provides a direct assessment of the calibration matrix’s accuracy at the given optimization iteration. In Equation~ \ref{eq:objective_function_evaluation}, $N$ represents the total number of landmarks, and \( \mathbf{p}_{\text{Phantom}, i} \) and \( \mathbf{p}_{\text{transformed}, i} = \mathbf{R}_{\text{ICP}} \cdot \mathbf{p}_{\text{world, i}}\) are respectively the coordinates of the \(i\)-th ground truth and transformed ultrasound landmarks.
\begin{equation}
    E_{\text{MSE}} = \frac{1}{N} \sum_{i=1}^{N} \|\mathbf{p}_{\text{transformed}, i} - \mathbf{p}_{\text{Phantom},i}\|^2
    \label{eq:objective_function_evaluation}
\end{equation}
%\[
%E_{\text{RMSE}} = \sqrt{\frac{1}{N} \sum_{i=1}^{N} \|\mathbf{p}_{\text{GT}, i} - %\mathbf{R}_{\text{ICP}} \cdot \mathbf{T}_{\text{tracking}} \cdot \mathbf{C} \cdot %\mathbf{p}_{\text{ultrasound}}\|^2}
%\]
% Where \( N \) is the total number of points in the alignment, and \( \mathbf{p}_{\text{Phantom}, i} \) and \( \mathbf{p}_{\text{transformed}, i} \) are the coordinates of the \(i\)-th ground truth point and transformed ultrasound point, respectively. The RMSE provides a measure of how well the ultrasound points match the ground truth after the transformation.

\textbf{Step 5: Optimization process}:
The optimization process is performed using gradient descent with the Adam optimizer, where the objective function defined in Equation  \ref{eq:objective_function_evaluation} is minimized over the seven parameters (three translations, three rotations and scaling) to determine the calibration matrix \( \mathbf{C} \). We use a learning rate $\alpha$ of \( 0.05 \) to balance convergence speed and optimization accuracy.

\section{Experiments and Results} \label{experiments}

\subsection{Data acquisition and implementation details}

The calibration algorithm was implemented in Python using PyTorch 2.0.1 and optimized with the Adam optimizer. All computations were performed on an NVIDIA RTX 4070 Ti GPU. US images were acquired using an ACUSON S3000 HELX Touch system with an IVUS (ACUSON AcuNav™ Ultrasound Catheter - 8 Fr / 90 cm) probe, covering depths from 7 to 11 cm. An EM tracking system (trakSTAR, NDI) with a 6-DoF sensor was attached to the IVUS near the transducer. For each depth, a separate acquisition was performed, where the probe was rotated clockwise for 10 seconds while the phantom remained fixed and fully immersed in water. The images (680×480 pixels, 15 fps) were captured using a frame grabber (AV.io HD+, Epiphan Video), yielding approximately 150 images per acquisition. The needle tips of the phantom (landmarks) were manually annotated and paired with their corresponding transformation matrices, with 15 landmarks recorded per acquisition. 

%For each IVUS depth, the calibration matrix was computed by minimizing the discrepancy between transformed image-space coordinates and known phantom needle positions, with convergence achieved in under one minute per depth. 

%\subsubsection{Calibration Dataset and Experiment}

\subsection{Calibration results}

For each acquired IVUS depth sequence, the calibration matrix is computed using the proposed approach. The optimization process converges in less than a minute for each sequence. The accuracy of the estimated calibration matrices is assessed by computing the RMSE between the landmark positions in the reconstructed US sequence and their corresponding positions in the 3D phantom. The results of our method are reported in Table~\ref{tab:calibration_results}, together with the results of the baseline PLUS method, which we also tested using the Perklab fCal-2.1 N-wire phantom \cite{lasso2014plus}. However, manipulating the IVUS transducer proved challenging for the baseline method, even for a surgeon experienced in handheld ultrasound calibration. Maintaining the phantom within the field of view while ensuring stable images for capturing target points/needles was not possible. This led to the baseline method failing to complete calibration at depths of $7$, $9$, and $11$ cm, with some attempts taking over an hour. In contrast, as shown in Table~ \ref{tab:calibration_results}, our method achieves results comparable to the baseline while maintaining a typical calibration error of 0.8–1.5 mm \cite{lasso2014plus}, which is on the same order of magnitude as the EM tracking system (1 mm), and being significantly faster. The overall time for our approach, including sequence acquisition, landmark annotation, and algorithm convergence, is under $10$ mins. Additionally, our method is better suited to the IVUS probe due to the phantom's design and is more adaptable to different US depths.

\begin{table}[ht]
\centering
\caption{Results of our method compared with the baseline PLUS method \cite{lasso2014plus} regarding accuracy, speed, and adaptability to varying depths.}\label{tab:calibration_results}
\begin{tabular*}{0.8\textwidth}{@{\extracolsep\fill}c c c c c }
\toprule
US Depth (cm) & \multicolumn{2}{c}{RMSE (mm)} & \multicolumn{2}{c}{Time (min)} \\
               & Ours & Baseline \cite{lasso2014plus} & Ours & Baseline \cite{lasso2014plus} \\
\toprule
$7$  & $1.16$ & $-$    & $ < 10 $ & - \\
$8$  & $0.88$ & $0.83$ & $ < 10 $ & $	\approx 60 $ \\
$9$  & $1.80$ & -    & $ < 10 $ & - \\
$10$ & $1.53$ & $1.12$ & $ < 10 $ & $	\approx 60 $ \\
$11$ & $1.79$ & -    & $ < 10 $ & - \\
\bottomrule
\end{tabular*}
\end{table}

\subsection{Preclinical \textit{in vivo} validation}

%\textbf{Animal model and data acquisition:} 

A porcine model with prior liver ablations and four tumors was used to simulate a clinically realistic liver ablation scenario. This setup presented challenges such as ultrasound shadows from air pockets and ablation zones, obscuring liver parts, representing a worst-case senario. IVUS acquisitions (single $360$° rotation scans) were performed at calibrated depths of $7$, $8$, $9$, $10$, and $11$~cm. However, only the $8$ and $9$ cm volumes provided sufficient anatomical detail for analysis. At $7$~cm, the field of view was limited, and at $10$ and $11$~cm, the resolution was inadequate for visualizing smaller vessels, as the key vascular structures were positioned outside of the optimal focal range. The reconstructed IVUS volumes at $8$ and $9$~cm were computed using the calibration matrix from the proposed approach and the 3D Slicer IGT \cite{lasso2014plus, fedorov20123d}, with an output spacing of $0.25$~mm/pixel. Additionally, a contrast-enhanced CT scan was obtained at the end of the experiment to serve as a high-resolution anatomical reference.

\begin{figure}[ht]
    \centering
    \begin{subfigure}{0.31\textwidth}
        \centering
        \includegraphics[width=\linewidth]{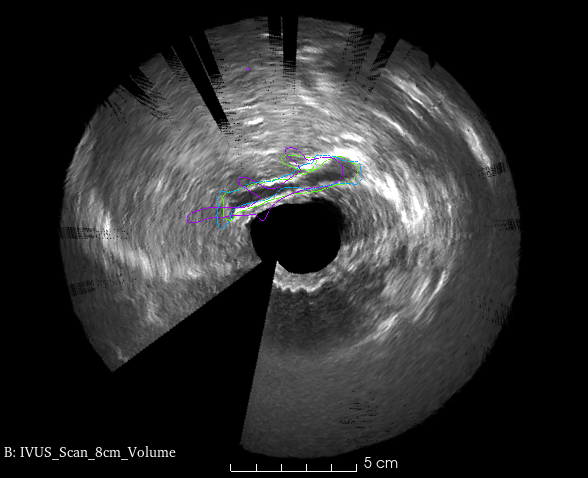}
        \caption{IVUS (axial view)}
    \end{subfigure}
    \begin{subfigure}{0.31\textwidth}
        \centering
        \includegraphics[width=\linewidth]{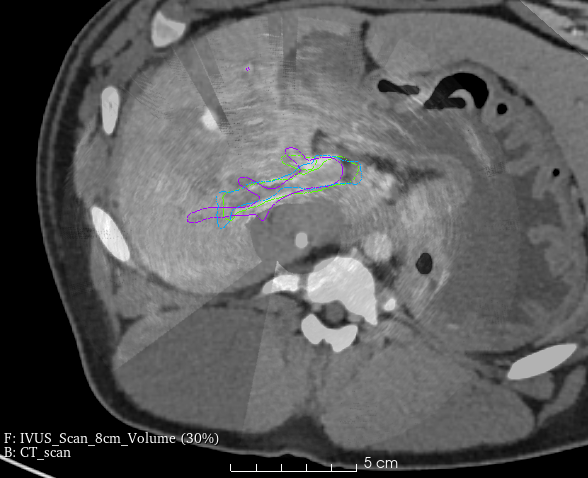}
        \caption{IVUS+CT (axial view)}
    \end{subfigure}
    \begin{subfigure}{0.31\textwidth}
        \centering
        \includegraphics[width=\linewidth]{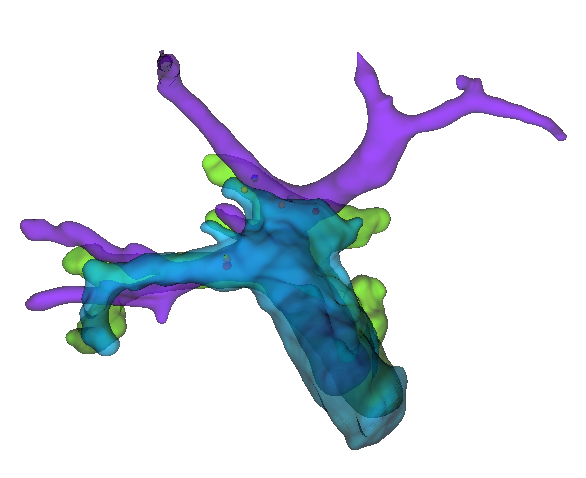}
        \caption{3D reconstruction}
    \end{subfigure}
    \caption{\textbf{Left}: IVUS volume in axial view; \textbf{Center}: IVUS overlaid with CT in axial view; \textbf{Right}: 3D reconstruction of the portal tree, based on the CT volume (purple), IVUS volume at 8~cm (green), and IVUS at 9~cm (blue)}
    \label{fig:ivus_ct_views}
\end{figure}

To evaluate our approach, the reconstructed IVUS volumes were segmented to extract the portal tree and then registered to the CT scan using liver surface landmarks, ensuring unbiased alignment and improving confidence in the spatial correspondence. The internal liver landmarks from both IVUS and CT volumes were compared, focusing on the Right Lateral Portal Vein (RLPV), Right Medial Portal Vein (RMPV), and Left Medial Portal Vein (LMPV). Ultrasound shadows from air pockets and ablation zones obstructed certain liver regions, preventing the identification of the Left Lateral Portal Vein. An expert surgeon independently identified bifurcations from both the IVUS volumes and the reference CT scan. Quantitative evaluation was performed by calculating the RMSE between identified vessel bifurcations in the IVUS and CT volumes, as shown in Table~\ref{tab:ivus_ct_rmse}. While our method yields an accurate reconstruction compared to the reference CT (see Fig~\ref{fig:ivus_ct_views}), some errors remain, which can be attributed, in part, to motion and deformation of the liver occurring between the IVUS and CT acquisitions. 
%Additionally, fixed-depth IVUS scanning has inherent limitations, especially since the portal tree spans multiple depths within the liver.
It is important to emphasize that our calibration performs optimally at specific depths and focal distances. However, the use of fixed-depth IVUS scanning limits validation, as key structures at varying depths within the liver fall outside the ideal focal ranges —particularly at US depths of 7, 10, and 11 cm—, leading to errors caused by imaging limitations.

\begin{table}[ht]
\caption{RMSE between IVUS reconstruction and reference CT for different anatomical landmarks.}\label{tab:ivus_ct_rmse}
\begin{tabular*}{\textwidth}{@{\extracolsep\fill}c c c c c}
\toprule
IVUS Depth (cm) & RLPV (mm) & RMPV (mm) & LMPV (mm) & RMSE (mm) \\
\toprule
8 & 1.35 & 7.14 & 6.70 & 5.71 $\pm$ 2.63 \\
9 & 1.82 & 2.88 & 4.78 & 3.40 $\pm$ 1.23 \\
\bottomrule
\end{tabular*}
\end{table}

\section{Conclusion and Discussion}\label{conclusion}
In this work, we present a fast and efficient calibration method for IVUS probes that enables accurate volume reconstruction. Using a custom-designed phantom and a gradient-based optimization algorithm, we estimate the calibration matrix between the image and tracking sensors. Compared to the PLUS method \cite{lasso2014plus}, our approach offers similar accuracy but is faster, easier to implement, and adaptable to different ultrasound depths.
Preclinical validation on a porcine liver demonstrated that our reconstructed IVUS volumes closely match the geometry of reference CT scans. We also achieved accurate alignment of internal anatomical landmarks. Optimal visualization of hepatic structures was observed at image depths of 8 and 9 cm, suggesting the potential of a multi-depth IVUS fusion approach to account for portal tree variability. Additionally, our findings emphasize the need for a depth-dependent calibration phantom with markers at the ultrasound image's focal distance to improve accuracy.
Future improvements include automating needle tip identification for fully automatic calibration, addressing sound speed variations, and developing depth-specific phantoms for enhanced accuracy. Our method also supports accurate 3D imaging, which could be useful for computer-aided interventions, including autonomous vessel tree identification \cite{beaudet2024towards}.

\begin{credits}
\subsubsection{\ackname} This work was partially supported by French state funds managed by the ANR under reference ANR-10-IAHU-02 (IHU Strasbourg).  The authors would like to express their gratitude to the Preclinical Research Unit of the IHU Strasbourg for their invaluable support and assistance throughout this study. 

\subsubsection{\discintname}

The authors have no competing interests to declare that are relevant to the content of this article.

\end{credits}


\bibliographystyle{splncs04}
\bibliography{bibliography}

\end{document}